\documentclass{ecai} 
\usepackage{latexsym}
\usepackage{amssymb}
\usepackage{amsmath}
\usepackage{amsthm}
\usepackage{booktabs}
\usepackage{enumitem}
\usepackage{graphicx}
\usepackage{color}
\usepackage{hyperref}
\usepackage[table]{xcolor}
\usepackage{pgfplotstable}
\usepackage{natbib} 
\newcommand{\BibTeX}{B\kern-.05em{\sc i\kern-.025em b}\kern-.08em\TeX}

\usepackage{color,soul}
\usepackage{caption}
\pgfplotsset{compat=newest}
\definecolor{darkblue}{rgb}{0, 0, 0.5}

\pgfplotstableset{
    color cells/.style={
        col sep=comma,
        string type,
        postproc cell content/.code={
            \ifnum\pgfplotstablerow=0
            \else
                \pgfkeysalso{@cell content=\rule{0cm}{2.4ex}\cellcolor{darkblue!##1}\pgfmathtruncatemacro\number{##1}\ifnum\number>10\color{white}\else \color{darkblue}\fi##1}%
            \fi
        },
        columns/x/.style={
            column name={},
            postproc cell content/.code={}
        }
    },
    every head row/.style={output empty row}, 
}
\usepackage{listings}
\usepackage{array}
\newcolumntype{C}[1]{>{\centering\arraybackslash}m{#1}}
\usepackage{multirow}
\usepackage{url}
\usepackage{todonotes}

\begin{document}

%%%%%%%%%%%%%%%%%%%%%%%%%%%%%%%%%%%%%%%%%%%%%%%%%%%%%%%%%%%%%%%%%%%%%%%%

\begin{frontmatter}

%%% Use this command to specify your submission number.
%%% In doubleblind mode, it will be printed on the first page.

\paperid{962} 

%%% Use this command to specify the title of your paper.

\title{Transformers in the Service of Description Logic-based Contexts}

%%% Use this combination of commands to specify all authors of your 
%%% paper. Use \fnms{} and \snm{} to indicate everyone's first names 
%%% and surname. This will help the publisher with indexing the 
%%% proceedings. Please use a reasonable approximation in case your 
%%% name does not neatly split into "first names" and "surname".
%%% Specifying your ORCID digital identifier is optional. 
%%% Use the \thanks{} command to indicate one or more corresponding 
%%% authors and their email address(es). If so desired, you can specify
%%% author contributions using the \footnote{} command.
        
\author[A]{\fnms{Angelos}~\snm{Poulis}}
\author[A]{\fnms{Eleni}~\snm{Tsalapati}}
\author[A]{\fnms{Manolis}~\snm{Koubarakis}} 

\address[A]{Dept. of Informatics and Telecommunications, \\ 
        National and Kapodistrian University of Athens \\ 
        \texttt{\{sdi1900230, etsalapati, koubarak\}@di.uoa.gr}}

%%% Use this environment to include an abstract of your paper.

\begin{abstract}
Recent advancements in transformer-based models have initiated research interests in investigating their ability to learn to perform reasoning tasks. However, most of the contexts used for this purpose are in practice very simple: generated from short (fragments of) first-order logic sentences with only a few logical operators and quantifiers. 
In this work, we construct the natural language dataset, DELTA$_D$, using the description logic language $\mathcal{ALCQ}$. DELTA$_D$ contains 384K examples, and increases in two dimensions: i) reasoning depth, and ii) linguistic complexity. In this way, we systematically investigate the reasoning ability of a supervised fine-tuned DeBERTa-based model and of two large language models (GPT-3.5, GPT-4) with few-shot prompting. 
Our results demonstrate that the DeBERTa-based model can master the reasoning task and that the performance of GPTs can improve significantly even when a small number of samples is provided (9 shots). We open-source our code and datasets.  %\footnote{Upon acceptance of paper}.
\end{abstract}

\end{frontmatter}

%%%%%%%%%%%%%%%%%%%%%%%%%%%%%%%%%%%%%%%%%%%%%%%%%%%%%%%%%%%%%%%%%%%%%%%%

\section{Introduction}
\begin{figure}
  \centering
  \includegraphics[scale=0.52]{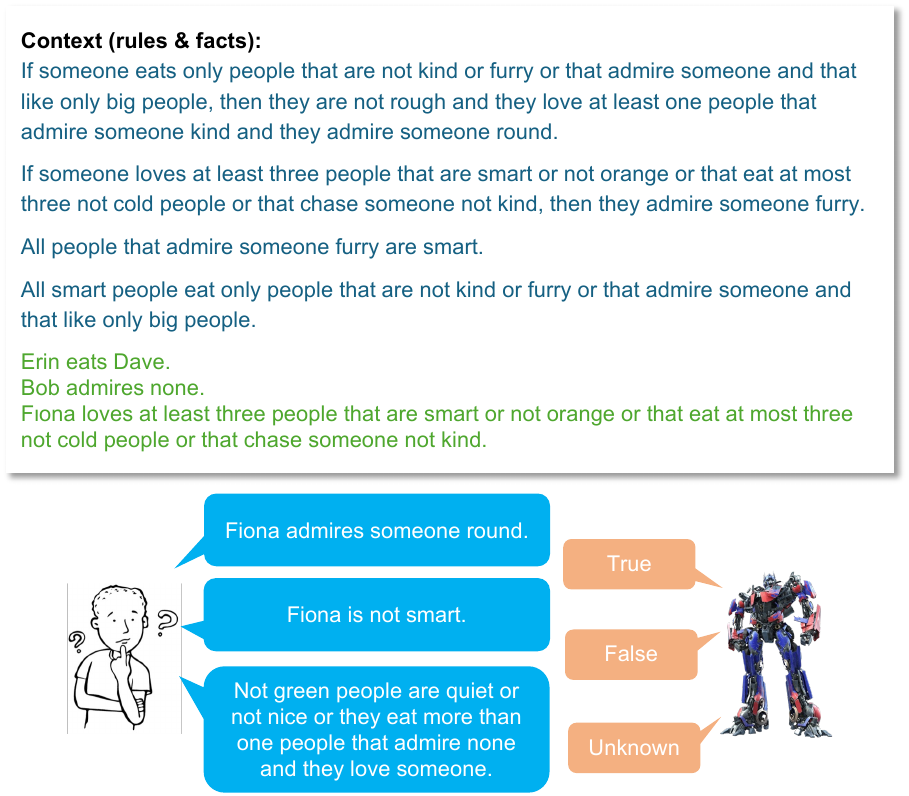}
  \caption[example]
  {An example from DELTA$_D$, where the context contains three sentences of high linguistic complexity level and the true and false sentences are of reasoning depth $3$.} 
  \label{fig:example}
\end{figure}

Description Logic (DL) languages~\citep{DBLP:conf/dlog/2003handbook} are fragments of first-order logic (FOL) that have evolved into one of the main formalisms for the representation of conceptual knowledge in a precise and well-defined manner. An expressive and decidable DL language that, besides the standard Boolean operators, supports existential, universal, and numerical constraints is $\mathcal{ALCQ}$. For instance,  in $\mathcal{ALCQ}$ one can formally express sentences like the ones appearing in Fig.~\ref{fig:example}.

The formal apparatus of DLs allows us to perform deductive reasoning tasks, such as \emph{entailment checking}, i.e., deciding whether a sentence or a set of sentences, logically implies another. With the latest advancements in transformer-based language models (TLMs) a new strand of research has emerged that investigates if TLMs can learn to carry out such tasks over \emph{contexts} expressed in natural language \citep{DBLP:conf/ijcai/ClarkTR20, folio, DBLP:journals/corr/abs-2305-14825, he-etal-2023-language}. However, in most cases, the contexts used were either composed of rather short sentences, simple in structure (i.e., their formal representations contain only a few logical operators and quantifiers)~\citep{DBLP:conf/ijcai/ClarkTR20, DBLP:conf/emnlp/TianLCX0J21, DBLP:conf/iclr/Saparov023}, or they were of limited size \citep{DBLP:conf/emnlp/TianLCX0J21, folio}. 

The fundamental question that this work seeks to answer is: \emph{``How well can TLMs perform inference over contexts produced from an expressive DL language, like $\mathcal{ALCQ}$?''}. A natural subsequent research question, in line with the literature ~\citep{DBLP:conf/ijcai/ClarkTR20, DBLP:conf/acl/TafjordDC21} but here is for a higher expressivity, is: \emph{``Is the performance of TLMs affected by the reasoning depth required to perform the inference process?''}. A third research question that arises is whether the fragment of the formal language is enough to measure the reasoning ability of a model. For instance, all sentences appearing in Fig.~\ref{fig:example} can be expressed formally within $\mathcal{ALCQ}$, but some are linguistically more complex than others, and one would expect that contexts containing mainly that complex sentences would be hard to process. Hence, the third  research question of this paper is: \emph{``Is the performance of TLMs affected by the linguistic complexity of the context?''}.

As discussed by \citet{madusanka-etal-2023-quantifiers}, the most suited reasoning problem to evaluate the impact of language constructs (like quantifiers and negation) is \emph{textual entailment checking}, i.e., entailment checking with natural language,  in a purely logical sense, eliminating the influence of any background or commonsense knowledge. Hence, to answer the above research questions, we have constructed the synthetic dataset DELTA$_D$ (DEscription Logics with TrAnsformers) of 384K examples (context-question-answer-depth-linguistic complexity) based on $\mathcal{ALCQ}$, where the question is the statement that we check whether it is logically deduced from the context, under the open world assumption. Besides the isolation of the commonsense/background knowledge, the synthetic nature of the dataset allows us to perform a systematic study of the performance of the TLMs, as DELTA$_D$ gradually increases both in reasoning and in linguistic complexity. Additionally, it allows us to eliminate obvious statistical features (e.g., the correlation between the answer ``False'' and the word ``not'' in the sentence in question).

We systematically tested the textual entailment checking ability of supervised fine-tuned DeBERTa and few-shot prompting on large language models (GPT-3.5, GPT-4) over DELTA$_D$. 
Our results show that the performance of the DeBERTa-based model, DELTA$_M$, is marginally affected when the reasoning depth increases or decreases (differently from \citep{DBLP:conf/emnlp/TianLCX0J21}) with average accuracy $97.5\%$. Also, we show that the length of the sentences does not affect the performance of the model (accuracy 99.7\% in max. reasoning depth and max. length of sentence). To ensure that DELTA$_M$ does not overfit on DELTA$_D$, and inspired by~\citep{DBLP:journals/corr/abs-2205-11502}, we changed the probability distributions used for the dataset generation and the accuracy of the model remained equally good. Additionally, testings to similar datasets~\citep{DBLP:conf/acl/TafjordDC21} returned good results. 

To check the impact of semantics on DELTA$_M$, we followed the approach of ~\citet{DBLP:journals/corr/abs-2305-14825} and we generated a symbolic ``translation'' of DELTA$_D$ by translating the pool of words used for the generation of the synthetic dataset to symbols. Zero-shot testings of DELTA$_M$ show that, in contrast to ~\citet{DBLP:journals/corr/abs-2305-14825}, it performs equally well, while GPTs' accuracy is slightly decreased. This indicates the performance of DELTA$_M$ is not affected by the semantics of the dataset (an expected result as the dataset is nonsensical). However, further translation of the dataset to resemble the language used to describe description logic sentences led to a significant reduction in the accuracy of the models. Finally, good testing results of DELTA$_M$ on a real-world scenario (fuel cell system diagnostics) demonstrate the potential of TLMs to be utilized in rule-based system diagnostics.

Overall, we make the following contributions:

($C_1$) We provide the first large, description logic benchmark of 384K examples. This is a significant contribution because building large benchmarks over expressive logic languages, like $\mathcal{ALCQ}$, is a challenging task as it requires performing query answering with logic reasoners, a process that can be very time-consuming ($\sim$ 1 min. for KB with long rules/facts of our dataset). Both the dataset and the code for its generation are openly available%\footnote{URL omitted due to double-blind review process but it can be found in the supplementary material.}.
\footnote{\url{https://github.com/angelosps/DELTA}}

($C_2$) We show that TLMs can perform entailment checking with very high accuracy over synthetic natural language contexts generated from $\mathcal{ALCQ}$ sentences. This demonstrates the potential of TLMs to be utilized for scalable reasoning tasks over vast KBs, thus bypassing formal representations required by traditional knowledge-based systems. 

($C_3$) We show that the performance of TLMs is not affected by the length of the sentences. 

($C_4$) We show that DeBERTa-based models are not affected by the vocabulary of the dataset. 

($C5$) We show how these contributions can be leveraged in a real-world use-case scenario.
\section{Background on Description Logics}
We can use $\mathcal{ALCQ}$~\citep{DBLP:conf/dlog/2003handbook} to represent knowledge about a domain by defining three types of entities: individuals (e.g., \emph{John}), concepts (e.g., \emph{Postdoc}, i.e., the concept describing the entities that are postdocs) and roles 
(e.g., \emph{teaches}). A \emph{concept expression} $C$ can be formed using these entities, Boolean constructors ($\sqcap$, $\sqcup$, $\neg$), quantifiers ($\forall$, $\exists$), and number restrictions ($\leq, \geq$) recursively as follows:
$C, D := A \mid \top \mid \bot \mid \neg C \mid C \sqcap D \mid C \sqcup D \mid \forall R. C \mid
\mbox{    }  \exists R. C \mid  \geq nR. C \mid  \leq nR. C $, where $A$ is an atomic concept, $R$ an atomic role, $\top$ the top concept, which has every individual as an instance, and $\bot$ the dual of $\top$.
In this way, one can represent formally complex concept expressions, such as all entities that ``\emph{have a Ph.D., teach at most two postgraduate courses and are not academics}'' ($\exists\textit{hasDegree.PhD} \sqcap \leq 2 \textit{teaches}. \textit{PostgrCourse}\sqcap \neg\textit{Academic}$). \emph{Rules} in $\mathcal{ALCQ}$ have the form $C \sqsubseteq D$ and describe relationships between concept expressions. For example, one can describe formally that all postdocs are described by the aforementioned concept as $\textit{Postdoc} \sqsubseteq \exists \textit{owns.PhD} \sqcap \leq 2 \textit{teaches}. \textit{PostgrCourse}\sqcap \neg Academic$. We denote with \emph{LHS} (left-hand side) the concept expression that appears on the left of the subsumption symbol ($\sqsubseteq$) in a rule and with \emph{RHS} (right-hand side) the concept expression that appears on the right. \emph{Facts} describe knowledge about named individuals, i.e., that are \emph{instances} of some concept (expression) and have the form $C(a)$ or $R(a, b)$, where $a$, $b$ individuals. Using complex expressions one can construct very complex facts. An $\mathcal{ALCQ}$ \emph{knowledge base} (KB) is a set of rules and a set of facts.

%For instance, one can represent formally all entities that ``\emph{have a Ph.D., teach at most two postgraduate courses and are not academics}'' ($\exists\textit{has.PhD} \sqcap \leq 2 \textit{teaches}. \textit{PostgrCourse}\sqcap \neg\textit{Academic}$). 
%\emph{Rules} in $\mathcal{ALCQ}$ describe relationships between concept expressions. For example, the fact that all postdocs are described by the aforementioned concept expression can be represented with the concept inclusion $\textit{Postdoc} \sqsubseteq \exists \textit{owns.PhD} \sqcap \leq 2 \textit{teaches}. \textit{PostgrCourse}\sqcap \neg Academic$. , e.g., John is a Postdoc ($\textit{Postdoc}(\textit{John})$), or how they related, e.g., John teaches Module05 ($\textit{teaches}(\textit{John},\textit{Module05})$).  
%The valid expressions in $\mathcal{ALCQ}$ are defined as follows. We start with a finite set of individuals, atomic concepts, and roles. If $A$ is an atomic concept and $R$ is a role,  according to the $\mathcal{ALCQ}$ grammar~\cite{DBLP:conf/dlog/2003handbook}, the
%concept expressions $C$, $D$ are constructed , i.e.,  with no individuals as instances. 

The \emph{inferred closure} of a KB $\mathcal{K}$ is the minimum set of rules and facts that can be logically inferred from $\mathcal{K}$. 
Given a KB $\mathcal{K}$ and a rule or a fact $a$, we say that $\mathcal{K}$ \emph{entails} $a$ (rule or fact) if every model  
of $\mathcal{K}$ (i.e., if every interpretation that satisfies all rules and facts of $\mathcal{K}$) is also a model of $a$. \emph{Entailment checking} can be considered as the prototypical reasoning task for querying knowledge: we check whether some statement is necessarily true, presuming the statements of the knowledge base. Following the semantics of DLs, we make the \emph{open-world assumption}, i.e., missing information is treated as unknown. 
Here, we consider $\textit{inferrence depth}$, or simply $\textit{depth}$ of $a$ with respect to $\mathcal{K}$, $\textit{depth}(a, \mathcal{K})$, as the size of the \emph{justification}~\citep{DBLP:conf/semweb/HorridgePS08} for $a$, i.e., the \emph{minimum} number of rules and facts in $\mathcal{K}$ that can be used to logically deduce that $a$ is true or false. If none of the two can be deduced, the answer is ``unknown'' and $a$ is not characterized by any depth.  

%In the following, to investigate transformers' performance with respect to the minimum number of inference steps a logical reasoner would require, we use the \emph{ justification}. This is a good approximation for two reasons: First, because it expresses the minimum depth of a proof. 
%and second because the probability to reuse the same axioms in the proving process is rather small as the KBs are small and they are generated in a chain-like as discussed in \ref{app:dataset generation}. 

%The \emph{justification} for an inferred axiom or fact from $\mathcal{K}$ is a
%minimal set of axioms from $\mathcal{K}$ that is sufficient for the entailment to hold.

\section{Dataset Generation}\label{sec:DatasetGen}
We investigate the ability of transformers to perform textual entailment checking over $\mathcal{ALCQ}$ KBs expressed in natural language with respect to two dimensions: i) the depth $\mathcal{D}$ of the sentences (i.e., rules/facts in question), henceforth mentioned as \emph{queries}, with respect to the corresponding KB, ii) the linguistic complexity level $\mathcal{L}$ (defined in Section~\ref{sec:KB generation}) of the knowledge required to answer the queries.
To achieve this, each \emph{example} in the dataset DELTA$_D$ is a 5-tuple $\langle \mathcal{T}$, $\mathcal{Q}$, $\mathcal{A}$, $\mathcal{D}$, $\mathcal{L} \rangle$,  where $\mathcal{T}$ is the \emph{context} containing $\mathcal{ALCQ}$ axioms (rules/facts) expressed in natural language, $\mathcal{Q}$ the query expressed in natural language, henceforth mentioned as \emph{question}, $\mathcal{A}$ is the \emph{answer} which can be either \emph{true}, \emph{false}, or \emph{unknown}, and $\mathcal{D}$ the depth of $\mathcal{Q}$, if $\mathcal{A}$ is \emph{true} or \emph{false}, otherwise it is denoted as $\texttt{na}$. $\mathcal{L}$ is the linguistic complexity of the KB.\footnote{DELTA$_D$ contains also the justification for each answer to be used for future work or by the research community for other downstream tasks, such as proof generation.} 

\begin{figure*}[t]
  \centering
  \includegraphics[scale=0.54]{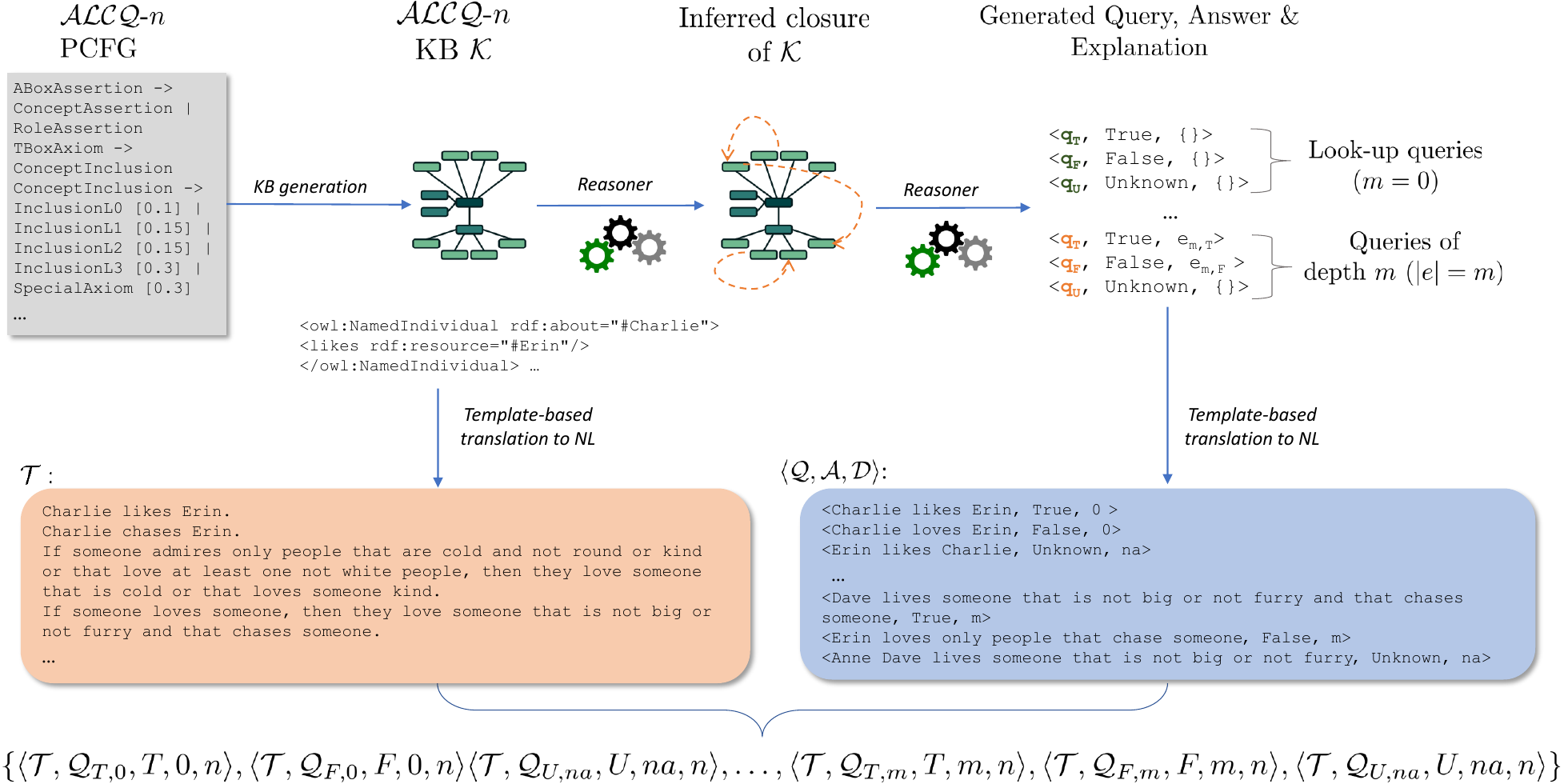}
  \caption[pipeline]
  {Data generation pipeline for examples with $n$-level context and answers of minimum inference depth $\leq m$}
 \label{fig:pipeline}
\end{figure*}

The pipeline for the generation of the dataset is presented in Figure~\ref{fig:pipeline}. For the generation of an example (described in detail in Section~\ref{sec:KB generation}) of linguistic complexity level $n$ ($\mathcal{L}\leq n$) and depth $m$ ($\mathcal{D}\leq m$), we first generate a KB $\mathcal{K}$ using a specially crafted probabilistic context-free grammar (denoted in Figure~\ref{fig:pipeline} with $\mathcal{ALCQ}$-$n$ PCFG) for producing rules and facts of maximum linguistic complexity $n$. Then, the inferred closure of $\mathcal{K}$ is calculated by utilizing the reasoner HermiT~\citep{DBLP:conf/ore/HorrocksMW12}, from which we calculate, as described in Section~\ref{sec:QueryGeneration}, \emph{true} (answer=true), \emph{false} (answer=false) and \emph{unknown} (answer=unknown) queries, which eventually will formulate the sentences in question. A KB is kept only if it can produce queries with all three types of answers at all depths up to $m$, otherwise a new one is generated. Once this process is completed, the generated queries (facts/rules) along with the original $\mathcal{K}$, are translated into natural language statements $\mathcal{Q}$ and into the context $\mathcal{T}$, respectively, by utilizing a set of natural language templates, as described in Section~\ref{sec:DL2NL}.

\subsection{KB Generation}\label{sec:KB generation}

To create diverse contexts and to avoid overfitting to a specific vocabulary, we have defined two different pools of terms, \emph{Pool A} and \emph{Pool B}. 
Pool A contains $14$ atomic concepts, $5$ roles, and $8$ individuals, mostly taken from RuleTaker dataset~\cite{DBLP:conf/ijcai/ClarkTR20} (in RuleTaker the rules are simple conjunctive implications, where the concept names are named ``attributes'', the roles ``relations'' and the individual names ``entities''). Pool B contains $8$ atomic concepts, $8$ roles, and $8$ individuals. Both pools can be found in Appendix~\ref{app:dataset generation}.

% \paragraph{Pool A}
% \begin{itemize}
%     \item \textbf{Atomic Concepts:} ``red'', ``blue'', ``green'', ``kind'', ``nice'', ``big'', ``cold'', ``young'', ``round'', ``rough'', ``orange'', ``smart'', ``quiet'', ``furry''.
%     \item \textbf{Role Names:} ``likes'', ``loves'', ``eats'', ``chases'', ``admires''.
%     \item \textbf{Individual Names:} ``Anne'', ``Bob'', ``Charlie'', ``Dave'', ``Erin'', ``Fiona'', ``Gary'', ``Harry''.
% \end{itemize}

From each pool, we generate
$20$ datasets ($40$ in total) of $1000$ KBs each, of various inference depths and axiom lengths. 

To obtain KBs of different linguistic complexity levels, we have manually crafted four types ($\mathcal{L} = 0$, $1$, $2$, $3$) of PCFGs, based on the number of constructors and quantifiers appearing in their axioms. %The definition of the various levels is provided in Appendix~\ref{app:dataset generation}. 
In general, a concept of linguistic complexity $\mathcal{L}$ contains $\mathcal{L}$ Boolean constructors and at most $\mathcal{L}+1$ quantifiers. 
Specifically, we define the \emph{linguistic complexity} level of a concept expression as follows: 
\begin{itemize}
    \item $Level_0$: 0  Boolean operators and $\leq$ 1 quantifiers. 
    \item $Level_1$: 1 Boolean operators and $\leq$ 2 quantifiers. 
    \item $Level_2$: 2 Boolean operators and $\leq$ 3 quantifiers. 
    \item $Level_3$: 3 Boolean operators and $\leq$ 4 quantifiers. 
\end{itemize}

An $\mathcal{L}$-type PCFG produces KBs of linguistic complexity level $\mathcal{L}$ with axioms that their one side (e.g., LHS) is of linguistic complexity $\mathcal{L}$ and their other side (e.g., RHS) of at most $\mathcal{L}-1$, but also contains simpler axioms, of smaller linguistic complexity levels. Specifically, A KB of level:  

%%% MOVED HERE FROM APPENDIX %%%

%\hl{Also, a rule that is generated by a PCFG of complexity level $\mathcal{L}$, is generated as follows:}

\begin{itemize}
\item $\mathcal{L}= 0$ contains axioms of the form $Level_0 \sqsubseteq Level_0 $,
\item $\mathcal{L}= 1$  contains axioms of the form $Level_0 \sqsubseteq Level_1 $,\\ $Level_1 \sqsubseteq Level_0 $, $Level_1 \sqsubseteq Level_1 $,
\item $\mathcal{L}=2$ contains axioms of the form  $Level_0 \sqsubseteq Level_2 $,\\ $Level_2 \sqsubseteq Level_0 $, $Level_1 \sqsubseteq Level_2 $, $Level_2 \sqsubseteq Level_1$,
\item $\mathcal{L}=3$ contains axioms of the form  $Level_0 \sqsubseteq Level_3 $, \\$Level_1 \sqsubseteq Level_3 $, $Level_2 \sqsubseteq Level_3 $, $Level_3 \sqsubseteq Level_0 $, $Level_3 \sqsubseteq Level_1 $, $Level_3 \sqsubseteq Level_2 $. 
\end{itemize}

For instance, KBs of level $\mathcal{L}=0$ contain only very simple facts or rules that do not contain any Boolean constructors but can contain one quantifier, such as $ \textit{Enthusiastic} \sqsubseteq \exists \textit{supports}. \textit{Enthusiastic}$ (translated in NL as ``Enthusiastic people support someone enthusiastic''), but KBs of level $\mathcal{L}=3$  can contain rules as complex as the first rule appearing in \ref{fig:example}.

It is important to discern the notion of linguistic complexity of a sentence from its length. We do not focus here only on sentences that contain, for instance,  multiple conjunctions but rather on sentences with a more complex structure (with quantifiers as well), leading to increased linguistic complexity. 

%KBs of level $\mathcal{L}=1$ contain rules that their LHS or RHS can contain up to a single Boolean constructor and up to two quantifiers, etc. 
To keep the KBs processible by the reasoners, the rules can contain up to seven atomic concepts and up to two nested quantifiers (e.g., $\exists \textit{likes}. (\exists \textit{loves}.\textit{Cat}))$, which describes the entities that like some entity that loves some cat). All KBs are rather small (with a minimum of 3 rules and 1 fact and a maximum of 14 rules and 12 facts) and are checked with respect to satisfiability and consistency with HermiT. 

\subsection{Query Generation}~\label{sec:QueryGeneration}
For an inference depth $\mathcal{D}$, a \emph{true query} $q$ is an axiom or fact selected from the inferred closure of a consistent $\mathcal{K}$, such that $\textit{depth}(q, \mathcal{K})= \mathcal{D}$. An \emph{unknown query} (answer=unknown) is generated by creating a random fact or statement (using the corresponding PCFG) such that it does not belong to the inferred closure of $\mathcal{K}$ and is consistent with $\mathcal{K}$. A \emph{false query} (answer=false) can be generated in three ways: 

\begin{itemize}
\item From an inconsistent $\mathcal{K}$: for every $a \in \mathcal{K}$ if $\mathcal{K} \setminus \{a\}$ is consistent then $a$ is a false query over the KB $\mathcal{K} \setminus \{a\}$.
\item From a consistent $\mathcal{K}$: i) By negating a true query $q$ with $\textit{depth}(q, \mathcal{K})= \mathcal{D}$ (and applying  De Morgan's laws). ii) By automatically generating an appropriate axiom or fact $a$ such that $\mathcal{K} \cup \{a\}$ is inconsistent and $\textit{depth}(a, \mathcal{K})=\mathcal{D}$. For instance, suppose that a KB $\mathcal{K}_1$ contains the axioms 
%\begin{gather*}
    $(\forall \textit{admires} .  \bot  ) ( Anne )$ and
    $\forall \textit{admires}. \bot  \sqsubseteq \forall \textit{likes} . \textit{Quiet}$
%\end{gather*}
which in natural language are translated into: ``Anne admires none'', ``All people that admire none like only quiet people''. Then, the fact $(\exists \textit{likes} . \neg \textit{Quiet})(\textit{Anne})$ stating that ``Anne likes someone who is not quiet'' forms a false query for $\mathcal{K}$. 
\end{itemize}

The disadvantage of the first approach is that it requires calling the reasoner multiple times, a time-consuming process, especially in KBs with long axioms (e.g., $\mathcal{L}$=3 KBs). Hence, we used the two latter approaches.

We set the reasoning depth limit to five (i.e., $\mathcal{D} = 0$, $1$, $2$, $3$, $5$) following the literature~\cite{DBLP:conf/ijcai/ClarkTR20}. Additionally, extending this further would require longer times for the dataset generation. 

%%% New from Appendix
%%% Is it ok to put this as a subsection? Or it should be a subsub section (under the Query Generation)?

%%%

\subsection{Data Translation to NL}~\label{sec:DL2NL}
The KBs and queries were translated to NL with the use of templates. The templates were created based on the user-friendly Manchester syntax for $\mathcal{ALCQ}$~\citep{DBLP:conf/owled/HorridgeDGRSW06}. Following this syntax, the intersection ($\sqcap$) and union ($\sqcup$) operators, are translated as ``and'' and ``or'', respectively, the existential ($\exists$) quantifier is translated as ``someone'' or ``something''  (depending on whether the pool is about people or things), the universal ($\forall$) as ``only'', and the number restrictions $\le, \ge$ as ``at most'' and ``at least''. Also, we use the word ``that'' for intersections and nested quantifiers. For instance, the fact $(\exists \textit{likes}.(\forall \textit{likes}. \textit{Kind}))(\textit{Bob})$ is translated as ``Bob likes someone that likes only kind people''.

Following the template-based approach suggested by \citet{DBLP:conf/acl/TafjordDC21}, the axioms of the form $C\sqsubseteq D$ are, roughly, translated into NL in four different ways: i)``If $C$ then $D$''; ii)``People/Things that are $C$ are $D$'', iii)``All people/things that are $C$ are $D$''; iv) If $C = \top $ and $D = \forall R . C'$ this is translated as ``Someone/something can $R$ only people/things that are $C'$''.
A fact $C(a)$ is translated as ``$a$ is $C$''.  
To ensure that the resulting NL sentences are grammatically correct we have used a grammar checker\footnote{\url{https://pypi.org/project/language-tool-python/}}.
\subsection{The Dataset DELTA$_D$}
At the end, the examples of the same depth and level from both pools are merged. This results in $20$ datasets of $2000$ KBs each, with each resulting dataset containing sentences from both vocabularies. From each KB we generated three queries (true, false, unknown) for each depth ($\mathcal{D}=\{0,1,2,3,5\}$), i.e., from each KB we generated $3 \times (d+1)$, $d\in \mathcal{D}$, queries. So, in total, the dataset contains $\Sigma_{d\in\mathcal{D}} 3 \times (d+1) \times 2000 \times (\mathcal{L}_{max} + 1)=384$K examples, as we generate KBs for each linguistic complexity level ranging from zero up to $\mathcal{L}_{max} = 3$. 

\subsection{Statistical Features}~\label{statfeatures}
As it is thoroughly discussed by \citet{DBLP:journals/corr/abs-2205-11502}, it is impossible to eliminate all statistical features that exist in data, besides, some of them inherently exist in logical reasoning problems. 

However, DELTA$_D$ is balanced with respect to some of the most obvious features: i) \emph{KB size}: From the same KB we extract all three types of questions (true, false, unknown); ii) \emph{Inference depth}: We keep a KB only if it can provide all three types of questions with the same inference depth; iii) \emph{Formulation of the question}: The translation to natural language is implemented in such a way that the word ``not'' appears almost equal number of times in true questions (52.39\%), false questions (50.71\%) and unknown questions (46.60\%); iv) \emph{Average length in words}: True questions 10.85, false questions 8.97, unknown questions 10.34.

%% FROM APPENDIX %%
\subsection{Usage of Long Axioms in the Proving Process}
\begin{figure}[h]
    \centering
    \includegraphics[width=1.0\linewidth]{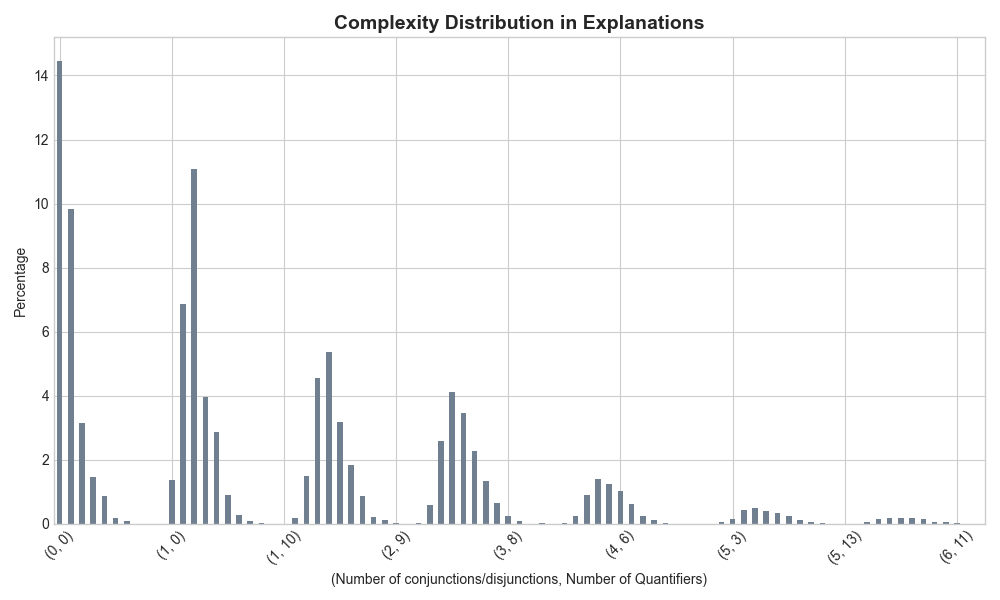}
    \caption{Complexity distribution analysis of explanation axioms in $\mathcal{L}=3$ KBs.}
    \label{fig:expl_express}
\end{figure}

Figure~\ref{fig:expl_express} shows the analysis of the linguistic complexity of facts and rules in explanations generated by $\mathcal{L}=3$ KBs. For each axiom, we measured the number of conjunctions/disjunctions and quantifiers. The horizontal axis shows each combination of conjunctions/disjunctions and quantifiers. The vertical axis shows the frequency (as a percentage) of each combination. We observe that more than 40\% of the axioms have at least two conjunctions/disjunctions \emph{and} at least two quantifiers. Also, it is worth noting that more than 67\% of the axioms have at least two conjunctions/disjunctions \emph{or} at least two quantifiers. This indicates that long axioms are necessary, and frequently used in the reasoning process for answering the questions.

% Hence, there is a high chance that the concepts in these explanations are of at least Level$_2$.

% We observe that only 10\% of the rules did not contain any quantifier (i.e., were of Level$_0$) and only 20\% did not contain any Boolean operator. 

% More than 50\% of the rules contained axioms with more than 2 Boolean operators and more than 60\% with more than 2 quantifiers. Hence, there is a high chance many of the rules in the explanation were of at least Level$_2$.} 
\section{Experiments}
\paragraph{Model Selection} We systematically tested the entailment checking ability of supervised fine-tuned DeBERTaV3-large, due to its recent advancements in NLU tasks~\citep{DBLP:conf/iclr/HeGC23}. We also tested in zero-shot and few-shot prompting the models GPT-3.5 (\texttt{gpt-3.5-turbo-1106}) and GPT-4 (\texttt{gpt-4}) from OpenAI, as they have demonstrated strong performance across various reasoning benchmarks \citep{DBLP:journals/corr/abs-2303-08774}. Our limited resources did not allow us to test the performance of other models, like the Llama family\footnote{\url{https://llama.meta.com/}}; we plan to do this in future work.   

\subsection{DeBERTa-based Models}

\subsubsection{Evaluation Setup} We fine-tuned the DeBERTaV3-large to predict true/false/unknown (i.e., multi-class sentence classification) for each example. A context-question pair was supplied to the model as \texttt{[CLS] context [SEP] question [SEP]}. We used accuracy as the evaluation metric. The test data has an equal balance of true/false/unknown answers, hence the baseline of random guessing is $33.3\%$.  
For our training, we used the AdamW optimizer \citep{loshchilov2019decoupled} using its default values for betas and a weight decay setting of $1e^{-4}$. The specifics of the chosen hyper-parameters, which we maintained consistently throughout our experiments, can be found in Appendix~\ref{app:hyper}. 

For each combination of depth and level ($5 \times 4 =20$ combinations in total), we trained different models on subsets of DELTA$_D$. A model DELTA$_{i,j}$ is trained in examples of reasoning depth up to $i$ and of linguistic complexity level up to $j$. For instance, the model DELTA$_{3,2}$ has been trained to depths up to $3$ and linguistic complexity levels up to $2$. The final model DELTA$_M$ has been trained to all depths and all linguistic complexity levels, i.e., DELTA$_M$=DELTA$_{5,3}$. 
For all datasets, we partitioned the data into 70\%/10\%/20\% splits for train/validation/test sets.

\begin{table}[t]
\caption{Accuracy of DELTA models on Test (own), on D$_{5,3}$ dataset, and slices of D$_{5,3}$ per depth.\\}
\centering
\resizebox{.95\linewidth}{!}{%
\pgfplotstabletypeset[color cells, col sep=&, row sep=\\]{
    x & x & x & x & x & x\\
   & DELTA$_{0,3}$ & DELTA$_{1,3}$ & DELTA$_{2,3}$ & DELTA$_{3,3}$ & DELTA$_{5,3}$\\ 
   Test (own) & 100.0 & 99.8 & 99.8 & 99.6 & 99.7 \\ 
   D$_{5,3}$ & 61.2 & 90.5 & 95.2 & 99.3 & 99.8 \\
   $\mathcal{D}$ = N/A & 100.0 & 99.4 & 99.5 & 99.2 & 99.7 \\
   $\mathcal{D}=0$ & 100.0 & 100.0 & 100.0 & 100.0 & 100.0 \\
   $\mathcal{D}=1$ & 43.4 & 100.0 & 100.0 & 100.0 & 100.0\\
   $\mathcal{D}=2$ & 24.5 & 73.1 & 99.5 & 100.0 & 100.0\\
   $\mathcal{D}=3$ & 34.1 & 71.3 & 99.5 & 100.0 & 100.0\\
   $\mathcal{D}=4$ & 29.2 & 77.6 & 84.5 & 99.5 & 100.0\\
   $\mathcal{D}=5$ & 19.3 & 76.5 & 83.5 & 98.5 & 99.5\\
}
}
\label{t:table1} 
\end{table}

\begin{table}[t]
\caption{Accuracy of DELTA models on Test (own) across all levels.}
\centering
\resizebox{.71\linewidth}{!}{%
\pgfplotstabletypeset[color cells, col sep=&, row sep=\\]{
x & x & x & x & x & x\\
 & $\mathcal{D}=0$ & $\mathcal{D}\leq1$ & $\mathcal{D}\leq2$ & $\mathcal{D}\leq3$ & $\mathcal{D}\leq5$ \\
   $\mathcal{L}=0$ & 100.0 & 99.7 & 99.4 & 98.9 & 98.9\\
   $\mathcal{L}\leq1$ & 100.0 & 99.7 & 99.6 & 99.7 & 99.5\\
   $\mathcal{L}\leq2$ & 99.9 & 99.5 & 99.7 & 99.7 & 99.6\\
   $\mathcal{L}\leq3$ & 100.0 & 99.8 & 99.8 & 99.6 & 99.7\\
}}
\label{t:testonOWN} 
\end{table}

\subsubsection{Evaluation Results} 
Table~\ref{t:table1} illustrates the performance of DELTA models when trained on up to $\mathcal{L}\leq 3$ linguistic complexity over the various inference depths (the results for smaller levels are presented in Appendix~\ref{app:performance_orig}). For instance, the column DELTA$_{0,3}$ shows the performance of the model trained on all levels in depth 0. Test (own) represents the (held out) test set of the dataset that the model has been trained on. The D$_{5,3}$ dataset has questions from all inference depths ($\mathcal{D}\leq5$) of all levels ($\mathcal{L}\leq3$). ``Depth N/A'' refers to the unknown questions, as these are not provable. ``$\mathcal{D}=0$'' to ``$\mathcal{D}=5$'' lines represent subsets of D$_{5,3}$ of 0-reasoning depth to 5-reasoning depth, respectively. 

It is observed that models trained even in $\mathcal{D}\leq 2$ datasets generalize quite well in larger depths (83.5\% for questions of $\mathcal{D}= 5$), while when trained in $\mathcal{D}\leq 3$ datasets they show impressive generalization ability (98.5\% for questions of $\mathcal{D}= 5$). Finally, we observe that the model trained in $\mathcal{D}\leq 5$ datasets almost masters (99.5-100\%) the reasoning task for all reasoning depths and linguistic complexity levels.

Table~\ref{t:testonOWN} demonstrates the performance of each model DELTA$_{i,j}$ when tested on Test (own). For instance, the cell that corresponds to $\mathcal{D}=0$, $\mathcal{L}=0$ shows the accuracy of the model DELTA$_{0,0}$. 
We observe that for all depths $\mathcal{D}=0$ to $\mathcal{D}\leq 5$ the models are robust across levels, hence increasing lengths do not affect their performance. 

We, also, partitioned the dataset to the various depths, i.e., we extracted from DELTA$_D$ $5$ datasets which contain only data of depth $\mathcal{D}=i$ (of all levels) and not up to $i$. Additionally, we trained a model on a set of $3,200$ examples specifically at depth $3$ for all lengths ($\mathcal{L}\leq 3$). The accuracy when tested in questions of depth $3$ was $97.5\%$, it slightly dropped when tested in questions of smaller depths ($\mathcal{D}=1,2$) to $\sim 94.5\%$, except for the look-up questions ($\mathcal{D}=0$), where the accuracy reached $99.0\%$. The model showed even better performance ($\sim97.8\%$) in larger depths ($\mathcal{D}=4,5$). Differently from the findings of~\citet{DBLP:conf/emnlp/TianLCX0J21}, these results demonstrate the model's capacity for generalization across both lesser and greater reasoning depths than those encountered during training. 

\subsubsection{Zero-shot Performance of DELTA$_M$ on Other Distributions.}

\paragraph{Results for Tweaked Dataset.} 
We generated the new dataset DELTA$_T$ by changing the probability distributions of the PCFG for $\mathcal{L} = 3$ as follows: We increased the probability of the universal quantifier ($\forall$) from $0.33$ to $0.70$ and the probability of the disjunction ($\sqcup$) from $0.50$ to $0.80$.
DELTA$_T$ contains $1,200$ examples of up to reasoning depth $\mathcal{D}\leq 5$. This tweaking has resulted in sentences with 0.8/sentence disjunctions and 0.62/sentence universals. The accuracy of DELTA$_M$ on DELTA$_T$ was $100.0\%$ for both true and false questions, $98.9\%$ for unknown questions, and, $99.6\%$ overall. As it is evident, the model is robust according to this tweaked distribution.

% \begin{table}[h!]
% \caption{Accuracy (per label and as total average) of DELTA$_M$ on DELTA$_T$.}
% \centering
% \resizebox{0.45\linewidth}{!}{%
% \pgfplotstabletypeset[color cells, col sep=&, row sep=\\]{
%   x                 & x         \\
%                     & Accuracy  \\
% label=True          & 100.0     \\
% label=False         & 100.0     \\
% label=Unknown       & 98.9      \\
% Total Average       & 99.6      \\
% }
% }
% \label{t:deberta_tweaked}  
% \end{table}

\paragraph{Results for ProofWriter Dataset~\citep{DBLP:conf/acl/TafjordDC21}.} 
The reason for choosing this dataset is that it was generated in a similar way (using PCFGs) as DELTA$_D$, it is under the open-world assumption and, partly, we have used the same pool of terms (Pool A). DELTA$_M$ demonstrated very high accuracy in true questions ($95.2\%$) and false questions ($94.0\%$) but low accuracy ($50.8\%$) in unknown questions. On average the accuracy was $75.7\%$. The very high accuracy for true/false questions is a surprising result as although DELTA$_D$ and ProofWriter have many common types of rules/facts, ProofWriter contains also rules that involve individual names (e.g., ``If Bob is blue then Erin is red'') and negated role assertions (e.g., ``Bob does not like Erin''), which are not supported by $\mathcal{ALCQ}$ and therefore are not contained in the training set of DELTA$_M$. The low performance of DELTA$_M$ in unknown questions can be attributed to the different generation processes among the two datasets. According to the generation process described in \citet{DBLP:conf/acl/TafjordDC21} (the source code is not openly available), the unknown questions in ProofWriter contain terms that appear in the context, whereas, as described in Section~\ref{sec:QueryGeneration}, unknown questions in DELTA$_D$ are formulated by choosing random terms from the corresponding pool, hence they can be completely irrelevant to the context. Hence, we can assume that this is a statistical feature that DELTA$_M$ may have learned. 

% \begin{table}[h!]
% \caption{Accuracy (per label and as total average) of DELTA$_M$ on ProofWriter~\cite{DBLP:conf/acl/TafjordDC21} depth=5 test-set.}
% \centering
% \resizebox{0.45\linewidth}{!}{%
% \pgfplotstabletypeset[color cells, col sep=&, row sep=\\]{
%   x                 & x         \\
%                     & Accuracy  \\
% label=True          & 95.2      \\
% label=False         & 94.0      \\
% label=Unknown       & 50.8      \\
% Total Average       & 75.7      \\
% }
% }
% \label{t:deberta_proofwriter}  
% \end{table}

\paragraph{Results for Use Case Scenario.}
We utilized the ontology rules and facts (generated from lab experiments) from~\citep{DBLP:journals/eswa/TsalapatiJJJLDM21} and generated $1,500$ examples for fuel cell system diagnostics. The context contained rules of the form ``If a system is in a state that is described by a low voltage value that is a result of an observation made by some voltage sensor that is a reliable sensor then the system is under some flooding'' and facts of the form ``v1 is a high voltage value''. A sample of this dataset is demonstrated in Appendix~\ref{app:fuel_cells} and the full dataset is also openly available. Again, DELTA$_M$ performed particularly well ($94.0\%$ accuracy). Detailed results of this section are presented in Appendix~\ref{app:fuel_cells}

\subsubsection{Handcrafted Quality Tests of DELTA$_D$}

To test the quality of DELTA$_D$ on which DELTA$_M$ is trained, we created simple test examples based on some of the most important knowledge base equivalences according to \citet{Rudolph2011}. The examples are demonstrated in detail in Appendix~\ref{app:qualityTest}. 

For instance, for the conjunction rule, we provided the context: ``Anne is red and green'' and the two (true) questions: ``Anne is red'' and ``Anne is green''. The model performs well overall, answering correctly $24/29$ questions, however, it seems that in contexts involving number restrictions, it returned the answer ``unknown'', which in two out of the six cases was wrong  
(notice though that the set of questions with number restrictions in DELTA$_D$ was balanced with respect to their answers).  

Additionally, it failed to learn the property $A \sqcap B \sqsubseteq A \sqcup B$.  
This became evident through the test: ``Context: Anne is red and green. Question: Anne is red or green. Answer: True. Prediction: Unknown''. To find where it fails in the reasoning chain, we asked the model the intermediate sentences ``Anne is green'' and ``Anne is red'', to which it returned (correctly) the answer ``true'', but it returned ``unknown'' to the question ``If someone is red and green then they are red or green''. Whereas, in the test ``Context: Anne is red. Anne is green. Question: Anne is red or green.'' the predicted answer was, again, falsely,  ``unknown''.  

\subsection{GPT Models}
\paragraph{Evaluation Setup} We tested GPT-3.5-turbo and GPT-4 models from the chat completion API provided by OpenAI. 

Our examples were limited to linguistic complexity $\mathcal{L}\leq1$, due to the models' context width limit: contexts of $\mathcal{L}\geq 2$ could not fit in the window for the few-shot setting. For the same reason, the maximum number of training shots was limited to $9$. To enforce deterministic behavior to the models we set \texttt{temperature}=$0$. To make the responses less verbose we set \texttt{max\_tokens}=$3$.

As transformer-based language models undergo pre-training through a certain form of language modeling objective, the most common approach to evaluate these models in the zero/few-shot setting is by employing prompt engineering techniques.  

To formulate our prompt, we used the  guidelines\footnote{\url{https://platform.openai.com/docs/guides/prompt-engineering/strategy-write-clear-instructions}} from OpenAI and our approach was based on the deductive reasoning prompts presented in~\citep{DBLP:journals/corr/abs-2305-14825}. The prompt that we concluded in was the following: \{``role'': ``system'', ``content'': ``You are an assistant capable of logical reasoning. Answer to the given question with `True', `False', or `Unknown' based on the context.''\}. 

\paragraph{Evaluation Results} In Table~\ref{t:gpt_delta} we present the average accuracy (over 100 examples) per inference depth of 0-shot and 9-shot prompting for GPT-3.5 and GPT-4. It is noted that GPT-4 has good performance (max 92\% and min 77\%) with just 9 shots. Also, it is evident that the models consistently struggle in increased inference depths, demonstrating that our dataset is challenging even for $\mathcal{L}\leq 1$.

\begin{table}[t]
\caption{Average accuracy per inference depth of 0-shot, 9-shot GPT-3.5, GPT-4 on 100 examples from DELTA of linguistic complexity level $\mathcal{L}\leq 1$.}
\centering
\resizebox{.95\linewidth}{!}{%
\pgfplotstabletypeset[color cells, col sep=&, row sep=\\]{
  x & x & x & x & x\\
 &0-shot GPT-3.5& 0-shot GPT-4 &9-shot GPT-3.5& 9-shot GPT-4\\
$\mathcal{D}$ = N/A & 60   & 98 & 75 & 92\\
$\mathcal{D} = 0$   & 74   & 80 & 84 & 89\\
$\mathcal{D} = 1$   & 68   & 57 & 82 & 86\\
$\mathcal{D} = 2$   & 48   & 42 & 60 & 77\\
$\mathcal{D} = 3$   & 43   & 32 & 67 & 80\\
$\mathcal{D} = 4$   & 44   & 30 & 57 & 83\\
$\mathcal{D} = 5$   & 46   & 40 & 57 & 83\\
}}
\label{t:gpt_delta}  
\end{table}

\subsection{Tests on Symbolic Data} To test the effect of the semantics of the words on the performance of DELTA$_M$ we created the dataset SoftSymbolic. SoftSymbolic was generated by replacing consistently in the test set the words appearing in pools A and B with symbols. Specifically, all individuals (e.g., \texttt{Anna}) were replaced with an $a_i$ symbol (e.g., \texttt{$a_3$}), all classes (e.g., \texttt{smart}) with an $C_j$ symbol (e.g., \texttt{$C_2$}), and all roles (e.g., \texttt{likes}) with an $R_k$ symbol (e.g., \texttt{$R_5$}), where $i, j, k$ is some ID number. The average performance of DELTA$_M$ over the SoftSymbolic dataset was $95.3\%$, hence in contrast to \citet{DBLP:journals/corr/abs-2305-14825}, we conclude that DELTA$_M$ is not affected by the lack of semantics.

To check if the models can perform over purely logical examples, we generated the HardSymbolic dataset, which resulted from the SoftSymbolic by also utilizing the DL terminology: the word ``some'' (corresponding to the existential quantifier) was translated as ``exists'', the ``if \ldots then \ldots'' (corresponding to subsumption) as ``is subsumed by'', etc.%\footnote{Samples of both datasets are presented in Appendix~\ref{app:symbolic_translation}}.

The performance of DELTA$_M$ on both Soft and Hard Symbolic datasets is presented in Table~\ref{t:deberta_symbolic}. We observe that the average performance of DELTA$_M$ over the HardSymbolic dataset dropped to $58.5\%$. This is an expected result as the structure of the tested sentences was very different from the sentences in which DELTA$_M$ had been trained. We tested also the GPT models on (100 examples of) the HardSymbolic dataset where they showed similar performance to DELTA$_M$. The results are shown in Table~\ref{t:gpt_symbolic}. The average accuracy of GPT-3.5 0-shot was $57\%$, 9-shot $61\%$; and GPT-4 $20\%$, $65\%$, respectively. Hence, TLMs seem to struggle with purely logical datasets.

\begin{table}[h!]
\caption{Average accuracy per inference depth of DELTA$_M$ on SoftSymbolic and HardSymbolic datasets.}
\centering
\resizebox{0.4\linewidth}{!}{%
\pgfplotstabletypeset[color cells, col sep=&, row sep=\\]{
  x & x & x \\
 &Soft. & Hard.\\
$\mathcal{D}$ = N/A & 84.1  & 71.8 \\
$\mathcal{D} = 0$   & 100.0 & 58.3 \\
$\mathcal{D} = 1$   & 99.3  & 39.4 \\
$\mathcal{D} = 2$   & 96.3  & 45.5 \\
$\mathcal{D} = 3$   & 94.9  & 68.2 \\
$\mathcal{D} = 4$   & 97.0  & 66.6 \\
$\mathcal{D} = 5$   & 95.6  & 60.7 \\
}}
\label{t:deberta_symbolic}  
\end{table}

\begin{table}[h!]
\caption{Average accuracy per inference depth of 0-shot, 9-shot GPT-3.5, GPT-4 on 100 examples (per depth) of the HardSymbolic dataset.}
\centering
\resizebox{.95\linewidth}{!}{%
\pgfplotstabletypeset[color cells, col sep=&, row sep=\\]{
  x & x & x & x & x\\
 &0-shot GPT-3.5& 0-shot GPT-4 &9-shot GPT-3.5& 9-shot GPT-4\\
$\mathcal{D}$ = N/A & 26  & 100 & 70 & 86 \\
$\mathcal{D} = 0$   & 72  & 32 & 58 & 78  \\
$\mathcal{D} = 1$   & 70  & 2 & 38 & 64 \\
$\mathcal{D} = 2$   & 60  & 2 & 64 & 66 \\
$\mathcal{D} = 3$   & 56  & 4 & 64 & 54 \\
$\mathcal{D} = 4$   & 54  & 0 & 64 & 58 \\
$\mathcal{D} = 5$   & 60  & 2 & 70 & 46 \\
}}
\label{t:gpt_symbolic}  
\end{table}

%The average performance of DELTA$_M$ over the HardSymbolic dataset dropped to $58.5\%$. This is an expected result as the structure of the tested sentences was very different from the sentences in which DELTA$_M$ had been trained. We tested also the GPT models, which showed similar performance (over 100 examples of this dataset): the average accuracy of GPT-3.5 0-shot was $57\%$, 9-shot $61\%$; and GPT-4 $20\%$, $65\%$, respectively. Hence, TLMs seem to struggle with purely logical datasets.
 
Samples of each dataset are presented in Appendix~\ref{app:symbolic_translation}.% and the evaluation results are presented analytically in Appendix~\ref{app:perform_symbolic}.
\section{Related Work}\label{sec:relWork}
\begin{table*}[ht!]
\caption{The state-of-the-art benchmarks for deductive reasoning with transformers.}
\tiny
\centering
\resizebox{\linewidth}{!}{%
\begin{tabular}{lccccc}
& \textbf{\#Questions}  & \begin{tabular}[x]{@{}c@{}}\textbf{Avg size of}\\ \textbf{sentences}\end{tabular} & \begin{tabular}[x]{@{}c@{}}\textbf{Max size of}\\\textbf{sentences}\end{tabular}  & 
\begin{tabular}[x]{@{}c@{}}\textbf{Formal}\\ \textbf{Language}\end{tabular}&  \begin{tabular}[x]{@{}c@{}}\textbf{Generating}\\ \textbf{method} \end{tabular}\\
bAbI Task15 \citep{DBLP:journals/corr/WestonBCM15}                      & 2K    & 4.6           & 5     & Role  assertions & Synthetic                     \\
Ruletaker \citep{DBLP:conf/ijcai/ClarkTR20}                             & 520K  & 6.13          & 27    & Conj. Rules  & Synthetic                     \\
ProofWriter \citep{DBLP:conf/acl/TafjordDC21}                           & 500K  & 6.07          & 27    & Conj. Rules   & Synthetic                     \\
BirdsElectricity  \citep{DBLP:conf/acl/TafjordDC21}                     & 5K    & 12.85         & 22    & Conj. Rules   & Synthetic                     \\
ParaRules$_{C/O}$  \citep{DBLP:conf/acl/TafjordDC21}                    & 40K   & 11.85/11.78     & 37/37 & Conj. Rules  & Synthetic \& Manual                             \\
FOLIO \citep{folio}                                              & 1.4K  & 10.58         & 59    &  FOL$\setminus$ Num. Restr. & Synthetic \& Manual          \\
PrOntoQA \citep{DBLP:conf/iclr/Saparov023}                            & 5.8K  & 6.2           & 20    & FOL$\setminus$ Num. Restr.   & Synthetic                             \\
LogicNLI\citep{DBLP:conf/emnlp/TianLCX0J21}                             & 30K   & 8.72          & 25    & FOL$\setminus$ Num. Restr. & Synthetic                            \\
\hline
\begin{tabular}[x]{@{}c@{}}\textbf{DELTA}$_D$ \\($\mathcal{L}_0$/ $\mathcal{L}_1$/ $\mathcal{L}_2$/  $\mathcal{L}_3$) \end{tabular}

& \begin{tabular}[x]{@{}c@{}}96K each,\\ 384K (in total)\end{tabular}       & \begin{tabular}[x]{@{}c@{}} 6.28/9.71/ \\ \textbf{13.08}/\textbf{13.08} \end{tabular}   & \begin{tabular}[x]{@{}c@{}} 26/38/\\ 50/\textbf{62} \end{tabular} & $\mathcal{ALCQ}$ & Synthetic                
\end{tabular}
} 
~\label{t:benchamarks}
\end{table*}
Multiple surveys \citep{DBLP:journals/corr/abs-2303-12023, DBLP:conf/acl/0009C23, DBLP:journals/corr/abs-2303-14725} in the literature describe the most recent research developments on the use of transformers for reasoning tasks. One of the first datasets generated for this purpose was from \citet{DBLP:conf/ijcai/ClarkTR20} with RuleTaker, demonstrating the potential of transformers to perform logical question answering under CWA by training LLMs on synthetic datasets. However, their approach was limited to short expressions of simple conjunctive rules. \citet{DBLP:conf/acl/TafjordDC21}, generated the ProofWriter datasets (under CWA and OWA) and with a T5~\citep{DBLP:journals/jmlr/RaffelSRLNMZLL20}-based model fined-tuned on ProofWriter showed that LLMs can generate proofs with high accuracy (94.8\% for depth 5). The generation approach of the DELTA$_D$ was based on the approach for the generation of the datasets RuleTaker and ProofWriter, i.e., using PCFGs. However, DELTA$_D$ is different from these datasets as i) $\mathcal{ALCQ}$ is a much more expressive logic language hence we produced new PCFGs; ii) we have defined different PCFGs for each linguistic complexity level (which has not been done for any other dataset in the literature); iii) it is balanced regarding the aspects discussed in Section~\ref{statfeatures}.

In more expressive contexts, \citet{DBLP:journals/corr/abs-2203-15099} showed that LLMs perform well (up to 90.5\%) over contexts generated by propositional logic and a small subset of FOL. \citet{folio}, with the FOLIO dataset (1.4K), generated from FOL sentences -but without number restrictions-, tested the ability of various LLMs for the same reasoning task and concluded that  RoBERTa~\citep{DBLP:journals/corr/abs-1907-11692} performed best among all tested models (including GPT-3 and Codex) %~\citep{DBLP:journals/corr/abs-2107-03374}),
but still, the performance was low. \citet{DBLP:conf/emnlp/TianLCX0J21} introduced the much richer synthetic dataset LogicNLI (30K), under OWA for diagnosing LLMs' ability in FOL reasoning, showing that even their best-performing model does not learn to perform reasoning tasks and cannot generalize to different scenarios. \citet{DBLP:conf/emnlp/SchlegelPP22} generated a very simple dataset (containing a single conjunction) for satisfiability checking and showed that models that perform well on hard problems do not perform equally well on easier ones, concluding that transformers cannot learn the underlying reasoning rules rather than they tend to overfit to patterns in the generated data. Also, \citet{DBLP:journals/corr/abs-2205-11502}, \citet{DBLP:conf/emnlp/TianLCX0J21} achieved similar results. \citet{bang2023multitask} studied ChatGPT's~\citep{Liu_2023} deductive reasoning ability on bAbi task 16~\citep{DBLP:journals/corr/WestonBCM15} and EntailmentBank~\citep{Dalvi2021ExplainingAW}, performing merely well. In addition, differently from our results (where the performance decrease was small), \citet{DBLP:journals/corr/abs-2305-14825} showed that LLMs perform significantly better when using natural language instead of symbolic representations of logical facts and rules. 

Most of the aforementioned benchmarks are composed of short sentences. The ones that contain longer sentences (avg. 13 words/sentence) are small ($\leq$ 40K), while none of them have examples with numerical restrictions. This is better demonstrated with Table~\ref{t:benchamarks}, where we present the metrics of the datasets that are most relevant to DELTA$_{D}$ (Entailment Bank is omitted as it does not conform to a specific formal language). A work that is close to our research is that of \citet{he-etal-2023-language}, who tested the ability of LLMs, and specifically of RoBERTa, to perform natural language inference tasks over existing OWL2 ontologies (e.g., FoodOn, Shema.org). However, the task studied is different, specifically, in \citet{he-etal-2023-language}, given two concept expressions $C$ and $D$ the TLM is asked to infer if one entails/contradicts the other, while in this work TLMs decide if a sentence can be inferred from \emph{a set of} rules and facts, i.e., a KB.
%They showed, that \texttt{roberta-large} can reach high performance ($\sim 92\%$) with only a few shots (128). However, in our work, we make a more systematic analysis of the ability of both fine-tuned DeBERTaV3 and few-shot prompted GPT models to perform logical entailments, both regarding the linguistic complexity of the sentences and the reasoning depth. Also, although OWL2 is more expressive than $\mathcal{ALCQ}$, number constraints are not taken under consideration in their study. Finally, the work of \citet{he-etal-2023-language} is limited to rules, omitting facts. 

Relevant to our research is also the work of \citet{madusanka-etal-2023-quantifiers}, who investigated the effects of the various types of quantifiers on the performance of TLMs. As the generated dataset is not currently openly available it is hard to evaluate its complexity and hence its relevance to DELTA$_D$. It is worth noting, though, that they do not investigate systematically the aspects that we have focused on in this work (inference depth, linguistic complexity). 

\section{Conclusions and Future Work}
We generated the only large dataset (384K) in the literature that targets expressive DLs (namely, $\mathcal{ALCQ}$), that enjoys both high expressivity and high linguistic complexity and is publicly available for further understanding of the functionality of TLMs. We showed that our DeBERTa-based model, DELTA$_M$, can carry out entailment checking over expressive synthetic datasets with very high accuracy, regardless of the linguistic complexity of the context. Differently from recent results in the literature, we showed that our model \textit{has} learned to generalize on unseen reasoning depths, smaller or greater. Zero-shot tests showed that DELTA$_M$ is mostly robust to other distributions. Tests with the GPT family showed that GPT-4 can have significant performance with only a few shots. The high accuracy of zero-shot testings in a real-world scenario demonstrates the potential of TLMs for performing reasoning tasks bypassing the necessity for domain experts to be familiar with formal representations. 

Our qualitative tests revealed the need for the development of systematic evaluation techniques of synthetically generated datasets. Hence, this will be our next step in future work.  Furthermore, we plan to explore the upper limit of the expressivity of the logic language so that a transformer-based model will be able to perform reasoning tasks with high accuracy. 

Finally, we will expand our evaluation section with other state-of-the-art generative models.

\newpage
\bibliography{mybibfile}

\appendix
\section{Appendix}

The appendix consists of the following content:

\begin{itemize}
    \item \ref{app:dataset generation}: Dataset generation
    \item \ref{app:hyper}: Additional training details
    \item \ref{app:performance_orig}: Evaluation results on datasets for $\mathcal{L}_0$, $\mathcal{L}_1$, $\mathcal{L}_2$
    \item \ref{app:symbolic_translation}: Symbolic translation
    % \item \ref{app:perform_symbolic}: Evaluation results on symbolic datasets
    \item \ref{app:fuel_cells}: Evaluation results on fuel cells data
    \item \ref{app:qualityTest}: Simple quality tests
\end{itemize}

\subsection{Dataset Generation}~\label{app:dataset generation}
We generated our synthetic dataset DELTA$_D$, using four probabilistic context-free grammars. To ensure that we will produce datasets with inference depth $\mathcal{D}>0$, i.e., datasets resulted from some inferencing, we generated a new statement $C\sqsubseteq D$, if $C$ either appeared in some already generated fact or the RHS of some already generated statement.

\subsubsection{Probabilistic Context-Free Grammars}
Each grammar is based on some vocabulary of terms. Pool A and Pool B are defined next. 
\paragraph{Pool A}
\begin{itemize}
    \item \textbf{Atomic Concepts:} ``red'', ``blue'', ``green'', ``kind'', ``nice'', ``big'', ``cold'', ``young'', ``round'', ``rough'', ``orange'', ``smart'', ``quiet'', ``furry''.
    \item \textbf{Role Names:} ``likes'', ``loves'', ``eats'', ``chases'', ``admires''.
    \item \textbf{Individual Names:} ``Anne'', ``Bob'', ``Charlie'', ``Dave'', ``Erin'', ``Fiona'', ``Gary'', ``Harry''.
\end{itemize}

\paragraph{Pool B}
\begin{itemize}
    \item \textbf{Atomic Concepts:} ``ambitious'', ``confident'', ``creative'', ``determined'', ``enthusiastic'', ``innovative'', ``logical'', ``persevering''.
    \item \textbf{Role Names:} ``admires'', ``consults'', ``guides'', ``instructs'', ``leads'', ``mentors'', ``supervises'', ``supports''.
    \item \textbf{Individual Names:} ``Ioanna'', ``Dimitrios'', ``Eleni'', ``Maria'', ``Manolis'', ``Angelos'', ``Panos'', ``Anna''.
\end{itemize}

To generate the KBs, we employ a random sampling technique to select a subset of individuals, roles, and atomic Concepts from the pools mentioned above. An item from each pool has the same probability of being chosen. 

The probabilistic context-free grammar for the linguistic complexity level $\mathcal{L}=0$ is shown in Table~\ref{tab:pcfg0}, the rest can be found in the supplementary material files. The PCFG shown is for Pool B. The grammars for the Pool A are similar. The probabilities in the PCFGs were determined experimentally to generate appropriate KBs that would yield the desired inferences in the minimum amount of time.  

% The level of \emph{linguistic complexity} $\mathcal{L}$ of a concept is defined as follows:
% \begin{itemize}
%     \item Level$_0$: 0 connectives (i.e., Boolean operators) and $\leq$ 1 quantifiers 
%     \item Level$_1$: 1 connectives and $\leq$ 2 quantifiers 
%     \item Level$_2$: 2 connectives and $\leq$ 3 quantifiers 
%     \item Level$_3$: 3 connectives and $\leq$ 4 quantifiers 
% \end{itemize}

% Specifically, (as it shown in Table~\ref{tab:pcfg0} for the $\mathcal{L}=0$ PCFG) the rules of KBs of each complexity level $\mathcal{L}$ are defined as follows:

% \begin{itemize}
% \item If KB is of level $\mathcal{L}= 0$ : $Level_0 \sqsubseteq  Level_0 $
% \item If KB is of level $\mathcal{L}= 1$ : $Level_0 \sqsubseteq  Level_1 $, $Level_1 \sqsubseteq  Level_0 $, $Level_1 \sqsubseteq  Level_1 $
% \item If KB is of level $\mathcal{L}=2$ : $Level_0 \sqsubseteq  Level_2 $, $Level_2 \sqsubseteq  Level_0 $, $Level_1 \sqsubseteq  Level_2 $, $Level_2 \sqsubseteq  Level_1$
% \item If KB is of level $\mathcal{L}=3$ : $Level_0 \sqsubseteq  Level_3 $, $Level_1 \sqsubseteq  Level_3 $, $Level_2 \sqsubseteq  Level_3 $, $Level_3 \sqsubseteq  Level_0 $, $Level_3 \sqsubseteq  Level_1 $, $Level_3 \sqsubseteq  Level_2 $ 
% \end{itemize}

\subsubsection{KB Sizes}
We utilize randomized parameters to control the size of a KB, based on the target reasoning depth of the corresponding dataset.
The optimal (as we have found through experimentation) predefined ranges of the rules and facts per reasoning depth $\mathcal{D}$ are as follows:
\begin{itemize}
    \item For $\mathcal{D}= 0$: $|\textit{rules}|\in [3,8]$, $|\textit{facts}| \in [1,5]$
    \item For $\mathcal{D}= 1$: $|\textit{rules}|\in[3,8]$, $|\textit{facts}| \in [2,6]$
    \item For $\mathcal{D}= 2$: $|\textit{rules}|\in[3,8]$, $|\textit{facts}| \in [3,8]$
    \item For $\mathcal{D}=3$: $|\textit{rules}|\in[4,8]$, $|\textit{facts}| \in [5,10]$
    \item For $\mathcal{D}=5$: $|\textit{rules}|\in[6,14]$, $|\textit{facts}| \in [6,12]$
\end{itemize}

% \subsubsection{Depth of Inference}
% To investigate transformers' performance with respect to the minimum number of inference steps a logical reasoner would require, we use the \emph{size of the smallest justification}~\citep{DBLP:conf/semweb/HorridgePS08} for the answers to the queries. This is a good approximation for two reasons: First, because it expresses the minimum depth of a proof, and second because the probability to reuse the same axioms in the proving process is rather small as the KBs are small and they are generated in a chain-like as discussed in \ref{app:dataset generation}.  

% The \emph{justification} for an inferred axiom or fact from $\mathcal{K}$ is a
% minimal set of axioms from $\mathcal{K}$ that is sufficient for the entailment to hold. 

\subsection{Additional Training Details}~\label{app:hyper}
We used PyTorch 2.0 to set up our training and testing (inferencing). We use the \texttt{microsoft/deberta-v3-large} model from the transformers\footnote{\url{https://github.com/huggingface/transformers}} library, along with the accelerate\footnote{\url{https://github.com/huggingface/accelerate}} framework.

We trained the DeBERTaV3-large (304M parameters) model on two A100 GPUs. We used mixed precision (FP16) for our calculations to save memory and speed up the process. The specific set of hyper-parameters used for all our models' training is given in Table~\ref{table:7}. The model showed significant performance with this set of hyper-parameters, so there was no reason to proceed with any further hyper-parameter tuning, especially given our limited resources. The model output corresponds to the truth value $0$ for \texttt{False}, $1$ for \texttt{True}, and $2$ for \texttt{Unknown} labels.

\begin{table}[h!]
    \caption{Detailed specifications of the hyper-parameters used in DeBERTaV3-large training.}
    \centering
    % \resizebox{\linewidth}{!}{%
    \begin{tabular}{ll}
        \toprule
        \textbf{Hyper-parameter} & \textbf{Value} \\
        \midrule
        Batch size & 4 \\
        Accumulation steps & 2 (Effective Batch size = 8) \\
        Learning rate & \(2 \times 10^{-5}\) \\
        Warm-up ratio & 0.06 \\
        Epochs & 4 \\
        Mixed precision & FP16 \\
        Betas & (0.9, 0.999) \\
        Weight Decay & \(1 \times 10^{-4}\) \\
        Text Embedding Size & 512 (dimensions) \\
        \bottomrule
    \end{tabular}
    % }
    \label{table:7}
\end{table}

\subsection{Evaluation Results on Datasets for $\mathcal{L}_0, \mathcal{L}_1, \mathcal{L}_2$}~\label{app:performance_orig}
The performance of the intermediate models DELTA$_{i,j}$, for $i\in\{0,1,2,3,5\}$, $j\in\{0,1,2\}$ on their corresponding datasets (of $\mathcal{D} \leq i$ and $\mathcal{L} \leq j$) are illustrated in Tables~\ref{t:table8}, ~\ref{t:table9}, ~\ref{t:table10}. We observe that the pattern of the models' performance across various linguistic complexity levels is similar. However, as the models progress to higher linguistic complexity levels and, hence are trained on more data, the number of times they achieve perfect accuracy is increased. The models trained on $\mathcal{D} \geq 3$ show very good generalization on unseen reasoning depths, whereas the performance on unseen reasoning depths of the models trained on $\mathcal{D} \leq 2$ fluctuates across linguistic complexity levels. This can be attributed to the complexity difference among linguistic levels, affecting models' generalization.

\begin{table}[h!]
\caption{Accuracy of DELTA models on their own test sets, and the entire, and slices of $\mathcal{D}\leq5,\mathcal{L}=0$ dataset.}
\centering
\resizebox{.95\linewidth}{!}{%
\pgfplotstabletypeset[color cells, col sep=&, row sep=\\]{
    x & x & x & x & x & x\\
   & DELTA$_{0,0}$ & DELTA$_{1,0}$ & DELTA$_{2,0}$ & DELTA$_{3,0}$ & DELTA$_{5,0}$\\ 
   Test (own) & 100.0 & 99.7 & 99.4 & 98.9 & 98.7\\
   $\mathcal{D}\leq5,\mathcal{L}=0$& 61.4 & 75.3 & 93.2 & 97.7 & 98.8 \\
   $\mathcal{D}$ = N/A & 98.8 & 97.9 & 94.4 & 95.2 & 98.2 \\ 
   $\mathcal{D}=0$ & 99.9 & 100.0 & 100.0 & 99.5 & 100.0 \\
   $\mathcal{D}=1$ & 48.9 & 99.6 & 100.0 & 99.5 & 100.0 \\
   $\mathcal{D}=2$ & 11.9 & 47.1 & 96.3 & 99.0 & 99.0 \\ 
   $\mathcal{D}=3$ & 34.3 & 49.5 & 75.7 & 99.0 & 99.0 \\
   $\mathcal{D}=4$ & 32.1 & 45.0 & 72.0 & 99.0 & 99.0 \\
   $\mathcal{D}=5$ & 29.6 & 42.9 & 67.2 & 97.5 & 99.0 \\
}}
\label{t:table8} 
\end{table}

\begin{table}[h!]
\caption{Accuracy of DELTA models on their own test sets, and the entire, and slices of $\mathcal{D}\leq5,\mathcal{L}\leq1$ dataset.}
\centering
\resizebox{.95\linewidth}{!}{%
\pgfplotstabletypeset[color cells, col sep=&, row sep=\\]{
    x & x & x & x & x & x\\
   & DELTA$_{0,1}$ & DELTA$_{1,1}$ & DELTA$_{2,1}$ & DELTA$_{3,1}$ & DELTA$_{5,1}$\\ 
   Test (own) & 100.0 & 99.7 & 99.6 & 99.7 & 99.5 \\
   $\mathcal{D}\leq5,\mathcal{L}\leq1$ & 67.9 & 92.4 & 85.8	& 98.7 & 99.6 \\
   $\mathcal{D}$ = N/A & 99.0 & 98.1 & 99.3	& 98.8 & 99.7 \\ 
   $\mathcal{D}=0$ & 99.9 & 100.0 & 100.0 & 100.0 & 100.0 \\
   $\mathcal{D}=1$ & 52.5 & 99.3 & 99.5	& 100.0 & 100.0 \\
   $\mathcal{D}=2$ & 27.4 & 81.7 & 97.5	& 99.0 & 99.5 \\ 
   $\mathcal{D}=3$ & 47.9 & 79.5 & 61.5 & 99.0 & 99.5 \\
   $\mathcal{D}=4$ & 46.9 & 77.2 & 58.0 & 98.0 & 99.0 \\
   $\mathcal{D}=5$ & 39.4 & 66.0 & 53.5 & 96.5 & 99.0 \\
}}
\label{t:table9} 
\end{table}

\begin{table}[h!]
\caption{Accuracy of DELTA models on their own test sets, and the entire, and slices of $\mathcal{D}\leq5,\mathcal{L}\leq2$ dataset.}
\centering
\resizebox{.95\linewidth}{!}{%
\pgfplotstabletypeset[color cells, col sep=&, row sep=\\]{
   x & x & x & x & x & x\\
   & DELTA$_{0,2}$ & DELTA$_{1,2}$ & DELTA$_{2,2}$ & DELTA$_{3,2}$ & DELTA$_{5,2}$\\ 
   Test (own) &  99.9 & 99.5 & 99.7	& 99.7 & 99.6 \\
   $\mathcal{D}\leq5,\mathcal{L}\leq2$ & 55.1 & 89.7 & 85.3	& 99.1 & 99.7 \\
   $\mathcal{D}$ = N/A & 99.6 & 95.6 & 98.7 & 99.2 & 99.7 \\ 
   $\mathcal{D}=0$ & 99.9 & 100.0 & 100.0 & 100.0 & 100.0 \\
   $\mathcal{D}=1$ & 37.2 & 99.6 & 100.0 & 100.0 & 99.5 \\
   $\mathcal{D}=2$ & 16.1 & 81.7 & 98.5 & 100.0 & 99.0 \\ 
   $\mathcal{D}=3$ & 16.4 & 62.7 & 59.0 & 98.5 & 99.5 \\
   $\mathcal{D}=4$ & 17.6 & 63.1 & 62.0 & 99.5 & 100.0 \\
   $\mathcal{D}=5$ & 10.0 & 63.4 & 51.0 & 98.0 & 98.5 \\
}}
\label{t:table10} 
\end{table}

\subsection{Symbolic Translation}~\label{app:symbolic_translation}
Examples of translations from natural language to both soft and hard symbolic forms are presented in Table~\ref{tab:symbolic_translations}. We can observe that the SoftSymbolic dataset is a good generalization test for DELTA$_M$ as it eliminates any vocabulary influence on the model, whereas the HardSymbolic dataset offers a purely logical form of the sentences. The symbolic datasets resulted from symbolic translations of the $\mathcal{D} \leq 5, \mathcal{L} \leq 1$ test-set of DELTA$_D$.

\begin{table*}[h!]
    \caption[Symbolic translations]{Examples of sentences' translations to soft/hard symbolic forms.}
    \centering
    \begin{tabular}{m{5.4cm}m{5.4cm}m{5.4cm}}
        \toprule
        \textbf{Natural language form} & \textbf{Soft symbolic form} & \textbf{Hard symbolic form} \\ \midrule

        If someone mentors someone that is ambitious and that supervises less than one creative people, then they guide only people that are not persevering or that consult at most two confident people. &
        
        If someone R6 someone that is C1 and that R7 less than one C3 people, then they R3 only people that are not C8 or that R2 at most two C2 people. &
        
        exists R6 . ( ( + C1 ) and ( $< 1$ R7 . ( + C3 ) ) ) is subsumed by only R3 . ( ( not C8 ) or ( $\leq$ 2 R2 . ( + C2 ) ) ) \\ \hline

        Maria supports less than one people that are confident or not persevering. &
        a4 R8 less than one people that are C2 or not C8. &
        ( $< 1$ R8 . ( ( + C2 ) or ( not C8 ) ) ) ( a4 ) \\ \hline

        If someone is not confident, then they mentor someone that is ambitious and that supervises less than one creative people. &
        If someone is not C2, then they R6 someone that is C1 and that R7 less than one C3 people. &
        not C2 is subsumed by exists R6 . ( ( + C1 ) and ( $< 1$ R7 . ( + C3 ) ) ) \\ \bottomrule
    \end{tabular}
\label{tab:symbolic_translations}

\end{table*}

\begin{table*}
\caption[L0 grammar]{Probabilistic Context-Free Grammar for $\mathcal{L}=0$ KBs}
\centering
% \small
\begin{tabular}{lll}
$ABoxAssertion$ & $\rightarrow$ & $ConceptAssertion \, | \, RoleAssertion$ \\
$TBoxAxiom$ & $\rightarrow$ & $ConceptInclusion$ \\
$ConceptInclusion$ & $\rightarrow$ & $InclusionL0$ \\
$InclusionL0$ & $\rightarrow$ & $ConceptL0 \, \sqsubseteq \, ConceptL0 \, [0.6] \, | \, SpecialAxiom \, [0.4]$ \\
$SpecialAxiom$ & $\rightarrow$ & $'+' \, \top \, \sqsubseteq \, \forall \, RoleName \, '.' \, '(' \, Concept \, ')' \, |$ \\
& & $\exists \, RoleName \, '.' \, '(' \, '+' \, \top \, ')' \, \sqsubseteq \, Concept$ \\
$Concept$ & $\rightarrow$ & $ConceptL0$ \\
$ConceptL0$ & $\rightarrow$ & $ConceptNameOrRestriction$ \\
$ConceptNameOrRestriction$ & $\rightarrow$ & $Polarity \, ConceptName \, | \, RestrictionConcept$ \\
$RestrictionConcept$ & $\rightarrow$ & $RestrictionD0$ \\
$RestrictionD0$ & $\rightarrow$ & $Restriction \, RoleName \, '.' \, '(' \, Polarity \, ConceptName \, ')' \, |$ \\
& & $\exists \, RoleName \, '.' \, '(' \, '+' \, \top \, ')' \, | \, \forall \, RoleName \, '(' \, '+' \, \bot \, ')$' \\
$Restriction$ & $\rightarrow$ & $\forall \, | \, \exists \, | \, Symbol \, Number$ \\
$Symbol$ & $\rightarrow$ & $'>' \, | \, '\geq' \, | \, '<' \, | \, '\leq' \, | \, '='$ \\
$Number$ & $\rightarrow$ & $'1' \, | \, '2' \, | \, '3'$ \\
$ConceptName$ & $\rightarrow$ & $'ambitious' \, | \, 'confident' \, | \, 'creative' \, | \, 'determined' \, |$ \\
& & $'enthusiastic' \, | \, 'innovative' \, | \, 'logical' \, | \, 'persevering'$ \\
$RoleName$ & $\rightarrow$ & $'admires' \, | \, 'consults' \, | \, 'guides' \, | \, 'instructs' \, |$ \\
& & $'leads' \, | \, 'mentors' \, | \, 'supervises' \, | \, 'supports'$ \\
$IndividualName$ & $\rightarrow$ & $'Ioanna' \, | \, 'Dimitrios' \, | \, 'Eleni' \, | \, 'Maria' \, |$ \\
& & $'Manolis' \, | \, 'Angelos' \, | \, 'Panos' \, | \, 'Anna'$ \\
$RoleAssertion$ & $\rightarrow$ & $RoleName \, '(' \, IndividualName \, ',' \, IndividualName \, ')'$ \\
$ConceptAssertion$ & $\rightarrow$ & $'(' \, Concept \, ')' \, '(' \, IndividualName \, ')$' \\
$Polarity$ & $\rightarrow$ & $'+' \, | \, '\neg'$ \\
$Connective$ & $\rightarrow$ & $'\sqcup' \, | \, '\sqcap'$
\end{tabular}
\vspace{5pt}

\label{tab:pcfg0}
\end{table*}

\subsection{Evaluation Results on Fuel Cells Data}\label{app:fuel_cells}
A sample from the fuel cells diagnostics datasets is presented in Table~\ref{tab:fuel_cells_examples}. We generated two such datasets, one involving a single sensor, and the other involving two sensors, hence making the contexts more complex. We generated these datasets using simple random sampling over predefined pools of examples so they could result in true, false, and unknown questions. We can observe that the vocabulary and the format of these data are different from the dataset DELTA$_D$ where our model was trained, although it performs particularly well ($94\%$) zero-shot.

\begin{table*}[h!]
    \caption[Fuel cell use case data]{Examples of generated $<$context, question, answer$>$ triplets about fuel cells. The first two triplets involve one sensor (s1) in their context, while the last one involves two sensors (s1 and s2).}
    \centering
    \begin{tabular}{m{8.6cm}m{4.6cm}m{3cm}}
        \toprule
        \textbf{Context} & \textbf{Question} & \textbf{Answer} \\ \midrule

        s1 is a system. \newline
        s1 is in a state st1. \newline
        st1 is described by v1. \newline
        v1 is result of an observation o1. \newline
        failure mode is not a normal mode. \newline
        o1 is made by vs. \newline
        vs is a voltage sensor. \newline 
        if a system is in a state that is described by a very high voltage value that is result of an observation made by some voltage sensor that is a reliable sensor then the system is under catalyst dissolution. \newline
        vs is a reliable sensor. \newline
        carbon support corrosion is a failure mode. \newline
        v1 is a very high voltage value. &
    
        The system is under catalyst dissolution. & 
        
        True. \\ \hline
        
        s1 is a system. \newline
        s1 is in a state st1. \newline
        st1 is described by v1. \newline
        v1 is result of an observation o1. \newline
        failure mode is not a normal mode. \newline
        o1 is made by vs. \newline
        vs is a temperature sensor. \newline
        if a system is in a state that is described by a low cathode humidity value that is calculated by some air relative humidity sensor that is a reliable sensor and some temperature sensor that is a reliable sensor then the system is under dehydration. \newline
        vs is a reliable sensor. \newline
        catalyst dissolution is a failure mode. \newline
        v1 is a low cathode humidity value. & 
        The system is not under dehydration. &
        False. \\ \hline
        
        s1 is a system. \newline
        s1 is in a state st1. \newline
        st1 is described by v1. \newline
        failure mode is not a normal mode. \newline
        v1 is calculated by vs1 and vs2. \newline
        vs1 is a hydrogen mass sensor. \newline
        vs2 is a temperature sensor. \newline
        if a system is in a state that is described by a very large anode humidity change that is calculated by some hydrogen relative humidity sensor that is a reliable sensor and some temperature sensor that is a reliable sensor and is described by a large cathode humidity change then the system is under membrane mechanical stress. \newline
        vs1 is a reliable sensor. \newline
        vs2 is a reliable sensor. \newline
        catalyst dissolution is a failure mode. \newline
        v1 is a very large anode humidity change that is calculated by some hydrogen relative humidity sensor that is a reliable sensor and some temperature sensor that is a reliable sensor and is described by a large cathode humidity value. & 
        The system is under carbon support corrosion. &
        Unknown. \\ \bottomrule
    \end{tabular}
\label{tab:fuel_cells_examples}
\end{table*}

\subsection{Simple Quality Tests}\label{app:qualityTest}
The handcrafted quality tests we created, targeting various important knowledge base equivalences, along with the predictions of DELTA$_M$ on those are presented in Table~\ref{tab:quality_tests}. Although these tests are very similar in both structure and vocabulary with the dataset that DELTA$_M$ has been trained, we can observe a blind spot of the model on cases involving numerical restrictions, tending to answer ``unknown'' (U) on such questions.

\begin{table*}
\caption[Handcrafted quality tests]{Handcrafted quality tests for DELTA$_M$.}
\centering
\begin{tabular}{@{}p{0.36\textwidth}p{0.33\textwidth}cc@{}}
\toprule
\textbf{Context} & \textbf{Question} & \multicolumn{1}{l}{\textbf{Correct Answer}} & \multicolumn{1}{l}{\textbf{DELTA$_M$}} \\ \midrule
\multirow{2}{*}{Anne is red and green.} & Anne is red. & T & T \\
 & Anne is green. & T & T \\ \midrule
\begin{tabular}[c]{@{}l@{}}Anne is red.\\ Anne is green.\end{tabular} & Anne is red and green. & T & \textbf{U} \\ \midrule
\multirow{2}{*}{If someone is blue, then they are red and green.} & If someone is blue, then they are red. & T & T \\
 & If someone is blue, then they are green. & T & T \\ \midrule
\begin{tabular}[c]{@{}l@{}}If someone is blue, then they are red.\\ If someone is blue, then they are green.\end{tabular} & If someone is blue, then they are red and green. & T & T \\ \midrule
If someone is blue, then they are red and green. & If someone is blue, then they are red or green. & T & \textbf{U} \\ \midrule
\begin{tabular}[c]{@{}l@{}}Anne is red.\\ Anne is green.\end{tabular} & Anne is red or green. & T & \textbf{U} \\ \midrule
\multirow{2}{*}{Anne is red and green.} & Anne is red or green. & T & \textbf{U} \\
 & Anne is green or red. & T & \textbf{U} \\ \midrule
Anne is red or green. & Anne is green or red. & T & T \\ \midrule
\multirow{2}{*}{If someone is blue or red then they are green.} & If someone is blue, then they are green. & T & T \\
 & If someone is red, then they are green. & T & T \\ \midrule
\multirow{2}{*}{\begin{tabular}[c]{@{}l@{}}If someone is blue, then they are green. \\ If someone is red, then they are green.\end{tabular}} & If someone is blue or red, then they are green. & T & T \\
 & If someone is blue and red, then they are green. & T & T \\ \midrule
\multirow{3}{*}{People that eat someone red or green, are blue.} & People that eat someone red or eat someone green, are blue. & T & T \\
 & People that eat someone red or green, they eat someone red or eat someone green. & T & T \\
 & People that eat someone that is red or eat someone that is green they eat someone that is red or green. & T & T \\ \midrule
Blue people eat someone red or green. & People that are blue they eat someone that is red or they eat someone that is green. & T & T \\ \midrule
People that eat only people that are red or green are blue. & People that eat only people that are red or eat only people that are green, are blue. & T & T \\ \midrule
\multirow{4}{0.35\textwidth}{People that eat something are blue. \\Anne eats Bob.\\ Bob is green.\\} & Anne is blue. & T & T \\
 & Anne is green. & U & U \\ 
 & & & \\ 
 \midrule
\multirow{5}{0.35\textwidth}{People that eat something are blue.\\Anne eats Bob.\\ Bob is green.\\ If someone is blue, then they are not green.\\} & Anne is blue. & T & T \\
 & Anne is green. & F & F \\
 & & & \\
 & & & \\ \midrule
\begin{tabular}[c]{@{}l@{}}Someone can like only people that are nice.\\ Bob is not nice.\end{tabular} & Anne likes Bob. & F & \textbf{U} \\ \midrule
\begin{tabular}[c]{@{}l@{}}Someone can like only people that are nice. \\ Bob is nice.\end{tabular} & Anne likes Bob. & U & U \\ \midrule
\begin{tabular}[c]{@{}l@{}}Anne likes less than two people.\\ Anne likes Bob.\\ Anne likes John.\end{tabular} & Anne likes Alice. & F & \textbf{U} \\ \midrule
Anne likes Bob. & Anne likes none. & F & F \\ \midrule
\multirow{3}{*}{\begin{tabular}[c]{@{}l@{}}Anne likes Bob.\\ Anne likes John.\\ Anne likes Alice.\end{tabular}} & Anne likes more than two people. & T & \textbf{U} \\
 & Anne likes more than four people. & U & U \\
 & Anne does not like less than two people. & T & \textbf{U} \\ \midrule
Anne likes Bob. & Anne does not like Bob. & F & F \\ \bottomrule
\end{tabular}
\label{tab:quality_tests}
\end{table*}

\end{document}